\pdfoutput=1
\documentclass[11pt]{article}

\usepackage{emnlp2021}

\usepackage{times}
\usepackage{latexsym}
\usepackage{todonotes}
\usepackage{microtype}
\usepackage{parskip}
\usepackage{microtype}
\usepackage{bm}
\usepackage{subcaption}
\usepackage{amsmath}
\usepackage{xcolor}
\usepackage{cleveref}
\usepackage{todonotes}
\usepackage{multirow}
\usepackage{booktabs}
\usepackage{xpatch}
\usepackage{blindtext}
\usepackage{amssymb}
\usepackage{hyperref}
\usepackage[T1]{fontenc}
\usepackage[utf8]{inputenc}

\usepackage{microtype}

\title{Document-level Entity-based Extraction as Template Generation}

\author{Kung-Hsiang Huang\textsuperscript{\rm 1,2}~~~
    Sam Tang\textsuperscript{\rm 3}~~~
    Nanyun Peng\textsuperscript{\rm 1,3}\\
    \textsuperscript{\rm 1} Information Sciences Institute, University of Southern California\\
    \textsuperscript{\rm 2} Department of Computer Science, University of Illinois Urbana-Champaign\\
    \textsuperscript{\rm 3} Computer Science Department, University of California, Los Angeles \\
    {\tt khhuang3@illinois.edu} ~~~
    {\tt samtang1430@gmail.com} \\
    {\tt violetpeng@cs.ucla.edu}  \\}

\begin{document}
\maketitle
\newcommand{\SideNote}[2]{\todo[color=#1,size=\small]{#2}} 
\newcommand{\ssteeve}[1]{\SideNote{orange!40}{#1 --steeve}}
\newcommand{\sam}[1]{\SideNote{green!40}{#1 --sam}}
\newcommand{\violet}[1]{\SideNote{purple!40}{#1 --violet}}
\newcommand{\Steeve}[1]{{\color{orange}#1}}
\newcommand{\modelshort}[1]{\textsc{TempGen}}
\newcommand{\modellong}[1]{Cross-attention Guided Template Generation}
\newcommand{\perpind}[1]{\textsc{PerpInd}}
\newcommand{\perporg}[1]{\textsc{PerpOrg}}
\newcommand{\target}[1]{\textsc{Target}}
\newcommand{\victim}[1]{\textsc{Victim}}
\newcommand{\weapon}[1]{\textsc{Weapon}}
\newcommand{\dygiepp}[1]{\textsc{DyGIE++}}
\newcommand{\scirexpipeline}[1]{\textsc{SciREX-P}}
\newcommand{\topkcopy}[1]{\textsc{TopK Copy}}
\newlength{\Width}%
\newlength{\DepthReference}
\settodepth{\DepthReference}{g}
\newlength{\HeightReference}
\settoheight{\HeightReference}{T}
\newcommand{\MyColorBox}[2][red]%
{%
    \settowidth{\Width}{#2}%
    \colorbox{#1}%
    {%      
        \raisebox{-\DepthReference}%
        {%
                \parbox[b][\HeightReference+\DepthReference][c]{\Width}{\centering#2}%
        }%
    }%
}
\setlength{\fboxsep}{1pt}

\begin{abstract}
    Document-level entity-based extraction (EE), aiming at extracting entity-centric information such as entity roles and entity relations, is key to automatic knowledge acquisition from text corpora for various domains. 
    Most document-level EE systems build extractive models, which struggle to model long-term dependencies among entities at the document level. 
    To address this issue, we propose a generative framework for two document-level EE tasks: role-filler entity extraction (REE) and relation extraction (RE). 
     We first formulate them as a template generation problem, allowing models to efficiently capture cross-entity dependencies, exploit label semantics, and avoid the exponential computation complexity of identifying N-ary relations. A novel cross-attention guided copy mechanism, \topkcopy~, is incorporated into a pre-trained sequence-to-sequence model to enhance the capabilities of identifying key information in the input document. Experiments done on the MUC-4 and \textsc{SciREX} dataset show new state-of-the-art results on REE (+3.26\%), binary RE (+4.8\%), and 4-ary RE (+2.7\%) in F1 score \footnote{The source code is publicly available at \url{https://github.com/PlusLabNLP/TempGen}}. 
\end{abstract} 
\section{Introduction}

Document-level entity-based extraction (EE) are tasks that extract entity-centric information, such as entities and their relations, from unstructured text across multiple sentences. %\ssteeve{this seems to be specific to relation extraction.}
With the rise of big data in recent years, document-level EE is growing in importance with applications such as understanding clinical reports \cite{nye2020understanding}, extracting document-level events~\cite{huang2021document}, and building knowledge graphs from journals \cite{wu2020extracting}. In this work, we focus on two classic tasks of document-level EE: role-filler entity extraction (REE) and relation extraction (RE). %\sam{i love this EE abbreviation}
% \ssteeve{We need to cite these applications.}

\begin{figure}[t]
    \centering
    \includegraphics[width=.95\linewidth]{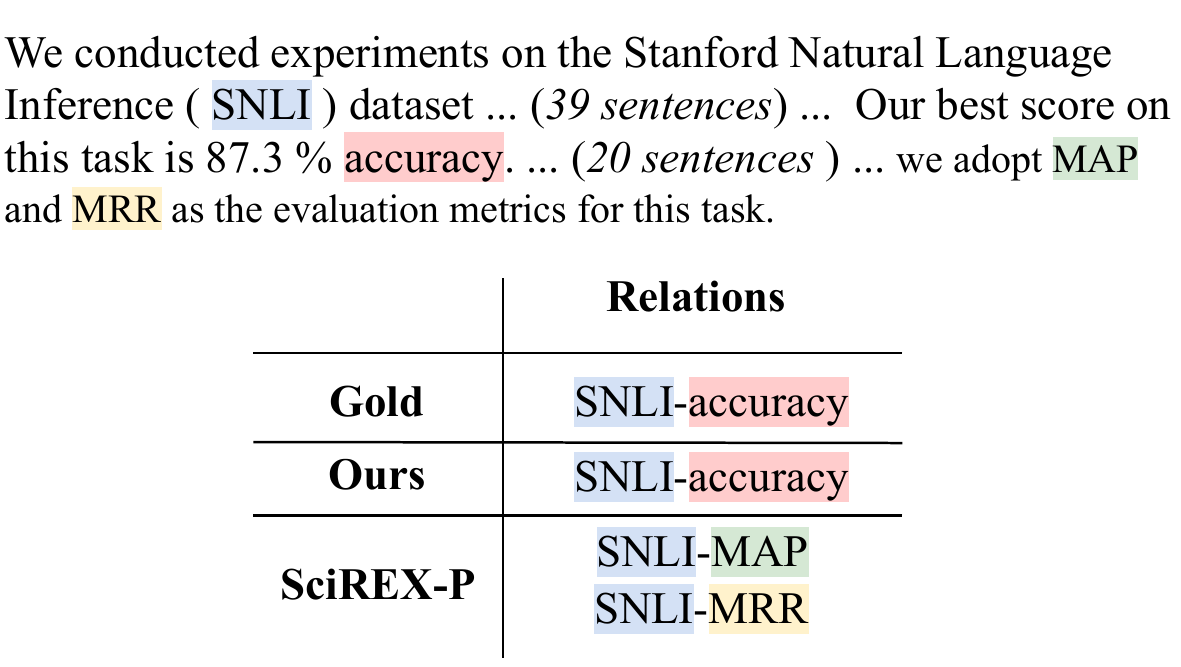}
    \caption{A comparison between our approach and a competitive extractive system, \scirexpipeline~ \cite{jain-etal-2020-scirex}, on a relation extraction example from \textsc{SciREX}. The task is to extract entities and identify which entities are related from the given scientific article. Due to the long distances between entities, \scirexpipeline~ struggles to extract the right entity pair that has a relation, while our approach correctly identifies them. This reflects our method's advantage in modeling long-term cross-entity dependencies.}
    % Given ``SNLI'', a \textsc{Material}-typed entity, the related entities in gold annotation are ``Natural Language Inference'' (\textsc{Task}) and ``accuracy'' (\textsc{Metric}). Even though the long distance between ``accuracy'' and ``SNLI'' (separated by 39 sentences) makes such relation hard to be identified, our approach successfully recognizes both \textsc{Task} and \textsc{Metric} entities. By contrast, \scirexpipeline~ outputs a great number of false positive \textsc{Task} entities and cannot identify the correct \textsc{Metric} entity.
    \vspace{-4mm}
    \label{fig:toy}
\end{figure}
% \violet{this figure should be self-contained such that even if people just jump right to this example, they can figure out what's going on here. For now, I have difficulties understand the task and the setup here. what does it mean by ``Given SNLP''?}
Recent works on document-level EE usually build task-specific classifiers on top of large pre-trained language models. For example, \citet{du-cardie-2020-document} builds a sequence tagging framework with multi-granularity representations based on \textsc{BERT} \cite{Devlin_2019} for role-filler entity extraction. \citet{jain-etal-2020-scirex} builds a relation extraction pipeline upon \textsc{SciBERT} \cite{beltagy-etal-2019-scibert}. However, there are a few drawbacks of this model architecture. First, as the size of the document increases, it becomes increasingly difficult for extractive methods to capture cross-entity dependencies among entitiy types due to long distances between entities, as shown in \Cref{fig:toy}. Additionally, discriminative models have no information regarding the semantics of the labels when classifying relations or entity types. Thus, it is unable to take advantage of the label semantics embedded in the pre-trained encoders.

Motivated by these challenges, we propose to formulate REE and RE tasks as template generation. Due to the autoregressive nature of generative setup, this formulation makes dependencies among the output entities easier to capture compared to sequence tagging methods. Moreover, label names are incorporated into the decoder targets for exploiting label semantics not present in the extractive counterparts. Furthermore, for tasks that involve the identification of $N$-ary relations, this formulation significantly alleviates the computational complexity of comparing exponential combinations of entities. A generative framework, \modellong~ (\modelshort~), that incorporates a novel copy mechanism into a pre-trained sequence-to-sequence model is proposed to solve the template generation problem effectively. %\violet{I'd first talk about the problem formulation, then model. here you basically only talked about formulation (although your original writing mentioned the model name, there were no details).}

Our contributions can be summarized as follows:
\begin{itemize}
  \item We propose to formulate document-level EE tasks as a template generation problem, which allows our generative framework to effectively capture cross-entity dependencies, better identify entities with label semantics, and avoid the exponential computation complexity of identifying $N$-ary relations.
  \item We devise a novel copy mechanism based on cross-attention to enable our model to better learn how to copy key information from the input document.
  \item Our approach achieves state-of-the-art results on MUC role-filler entity extraction task and \textsc{SciREX} relation extraction task, while being data efficient compared to previous systems.
\end{itemize}

\begin{figure*}[t]
    \centering
    \includegraphics[width=.95\linewidth]{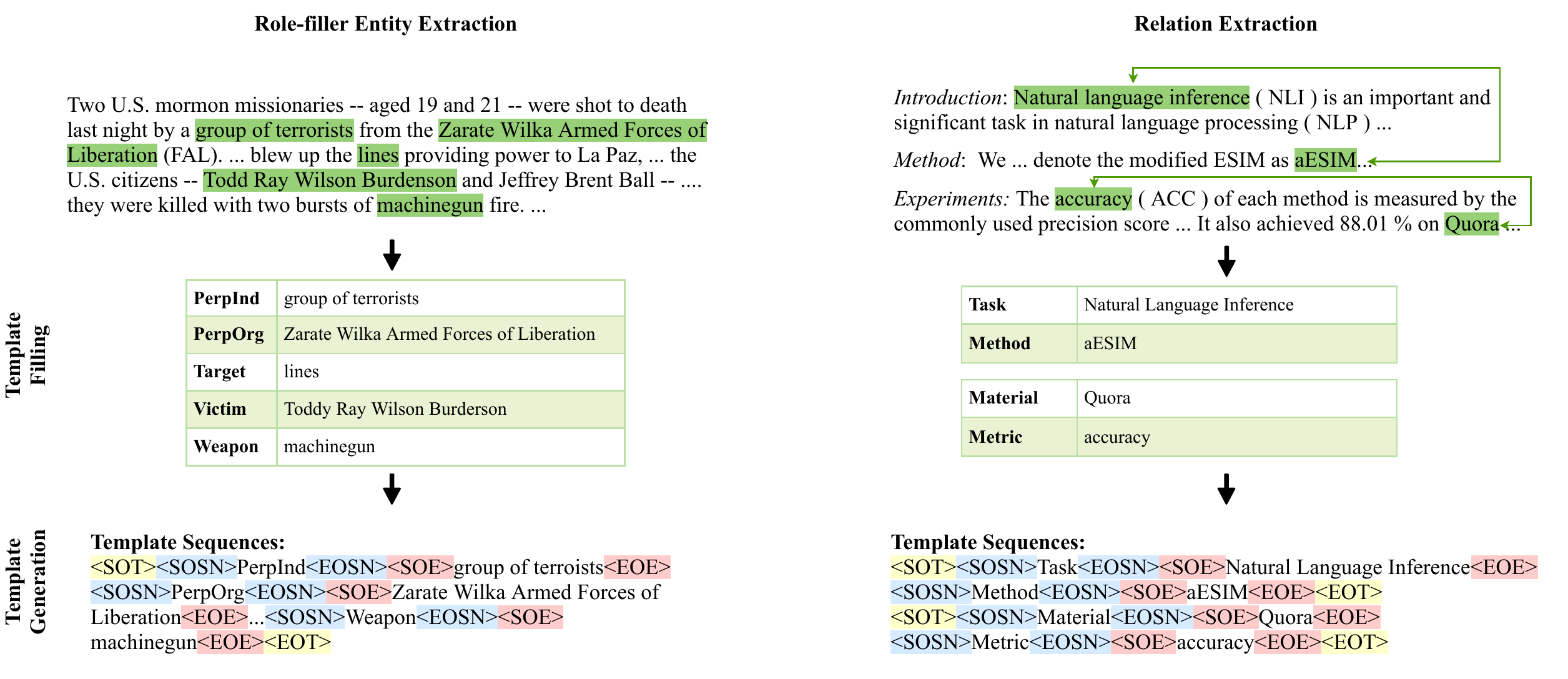}
    \caption{An overview of how document-level EE tasks can be transformed into template generation tasks. Special tags are defined as follows: \MyColorBox[yellow!10]{<SOT>}: start of template, \MyColorBox[yellow!10]{<EOT>}: end of template, \MyColorBox[cyan!10]{<SOSN>}: start of slot name, \MyColorBox[cyan!10]{<EOSN>}: end of slot name, \MyColorBox[red!10]{<SOE>}: start of entity (slot value), \MyColorBox[red!10]{<EOE>}: end of entity.}
    \label{fig:method}
    \vspace{-5mm}
\end{figure*}
\section{Tasks}
This section gives an overview of the two document-level EE tasks we tackled in this work: role-filler entity extraction (REE) and relation extraction (RE). %\sam{maybe good to add a sentence motivating why we choose these two tasks}
% \sam{maybe move figure 2 up closer to section 2?}
\subsection{Role-filler Entity Extraction}
The REE task aims to extract all entities involved in events from the input article \cite{du-2020-grit}. 
It differs from the event template extraction task introduced by the MUC-4 dataset \cite{muc-1992-message} in that only one event template, as opposed to many, is outputted for each input document. For documents associating with multiple event templates (all events in MUC-4 are of \textsc{Attack} type), the event templates are collapsed as one --- systems are required to identify all entities associate with different events for each role type. An event template consists of a set of pre-defined roles, and each role is filled with zero to many entities, as shown in \Cref{fig:method}. An entity is characterized by a group of mentions, which are spans of text in the input document. %Systems are only required to identify at least one correct mention for each entity. 

\subsection{Relation Extraction} %\violet{I think what's the input and the output should be clearer defined. e.g., do you assume entities are given or you need to identify? Do you expect output typed relation, or binary (yes/no) relations?}
We focus on end-to-end document-level relation extraction where systems first extract entities from the input document and then identify the $N$-ary non-typed relations among the extracted entities. The \textsc{SciREX} \cite{jain-etal-2020-scirex} dataset is the only dataset that supports such end-to-end configurations that we know of. Thus, we follow the definition of document-level RE in \textsc{SciREX}, which contains binary and 4-ary relation annotation. A binary relation contains two typed entities, and a 4-ary relation contains four typed entities. An entity is represented by a cluster of mentions, similar to the REE task. Systems should first extract salient entities of pre-defined types\footnote{Salient entities are entities needed to describe the results of corresponding scientific article.}. Then, binary and 4-ary relations among salient entities are identified. A binary relation example is shown in \Cref{fig:method}.
%Given an input document, entities of pre-defined types are first extracted, where an entity corresponds to a set cluster of text spans. Then, non-salient entities, , are filtered. Finally, binary and 4-ary relations among the predicted salient entities are identified.  \sam{there's inconsistency between this section and the previous one. If you discuss our generative formulation in RE, you should do the same inn REE} 
\section{Proposed Methods}
In this section, we first illustrate how the REE and RE tasks can be framed as a template generation problem. This formulation then allows us to capture cross-entity dependencies easily with our proposed generative model, a pre-trained sequence-to-sequence model integrated with a copy mechanism.

\subsection{Template Generation Formulation}
\label{sec:ie_as_tempgen}
We frame the REE and RE tasks as template generation problem, as shown in \Cref{fig:method}. A template is composed of slot names and slot values. For both tasks, slot names are entity types, and slot values are all entity mentions corresponding to such entity types. Similar to previous works on REE \cite{huang-riloff-2011-peeling, du-cardie-2020-document, du-2020-grit}, we only generate one template per document without differentiating which event template each entity mention associates with. In contrast, for RE, we generate multiple templates, each corresponding to a relation. A binary relation can be represented by a template of 2 slots, whereas a 4-ary relation forms a 4-slot template. A relation template consists of typed mentions of corresponding salient entities. After transforming REE and RE annotation to templates, each template can then be transformed into \textit{template sequences} with special tags delimiting templates, slot names, and slot values. %With our approach, the computation complexity of predicting relations among exponential combinations of entities is avoided.

Formally, a document of tokens $\mathcal{D} = \{\mathcal{D}_i\}^n_{i=1}$ may correspond to a decoding target of zero to many \textit{template sequences} $\{T_i\}^l_{i=1}$. A \textit{template sequence} $T_i$ is characterized by multiple \textit{slot sequences} $\{S_{i,j}\}^m_{j=1}$, 

{
\vspace{-4mm}
\begin{equation}
    \small
    \begin{gathered}
    \nonumber
     T_i = \MyColorBox[yellow!10]{<SOT>} \ S_{i,1}, ..., S_{i,m} \ \MyColorBox[yellow!10]{\text{<EOT>}}. \\
    \end{gathered}
\end{equation}
}A \textit{slot sequence} $S_{i,j}$ is represented by slot names and entities,

{
\vspace{-4mm}
\begin{equation}
    \small
    \begin{gathered}
    \nonumber
     S_{i,j} = \MyColorBox[cyan!10]{\text{<SOSN>}} \ L  \ \MyColorBox[cyan!10]{\text{<EOSN>}} \
            \MyColorBox[red!10]{\text{<SOE>}} \ \mathcal{D}^{(e_k)}_1 ... \mathcal{D}^{(e_k)}_n  \ \MyColorBox[red!10]{\text{<EOE>}}. \\
    \end{gathered}
\end{equation}
}where $L$ is the slot name\footnote{Slot name corresponds to role in REE and entity type in RE. }, and $\mathcal{D}^{(e_k)}_1, ..., \mathcal{D}^{(e_k)}_n$ is the token sequence that correspond to one mention randomly sampled from entity $e_k$. Special tokens, such as \MyColorBox[cyan!10]{<SOSN>} and \MyColorBox[cyan!10]{<EOSN>}, are to indicate whether a tag-enclosed string is a slot name or an entity mention. In the first row of the REE example from \Cref{fig:method}, $L$ would be \textsc{PerpInd} and $\mathcal{D}^{(e_k)}_1, ..., \mathcal{D}^{(e_k)}_n$ are ``group of terrorists''. Using this formulation, scalability challenges of modeling cross-entity dependencies is alleviated due to the significantly reduced distances between entities in \textit{template sequences}. %There are four main advantages of this formulation: (1) Scalability challenges of modeling cross-entity dependencies is alleviated due to the significantly reduced distance between entities in \textit{template sequences}. (2) Since later generated entities are conditioned on entities that have already been predicted, cross-entity dependencies can be implicitly captured. (3) The slot name semantics can be exploited for better entity identification. %(4) The exponential computation complexity of identifying n-ary relations is avoided.
%\sam{in general I think there can be more elaboration on this section}

\subsection{\modellong~}
\label{sec:model_detail}
The template generation problem can be broken down into two sub-goals: (1) generating valid template structures while capturing the dependencies between the input document and decoder targets, and (2) ensuring that salient mentions in the input document are correctly identified and outputted by the decoder. To achieve the first sub-goal, we leverage BART \cite{lewis-etal-2020-bart}, a pre-trained sequence-to-sequence model. The second sub-goal is achieved using a novel copy mechanism incorporated into BART.

\paragraph{Seq2Seq Model for Template Generation}
BART \cite{lewis-etal-2020-bart} is a pre-trained language model that combines bidirectional and auto-regressive transformers. Pre-training with multiple denoising objectives, BART has demonstrated significant advantages in various text generation tasks, especially on summarization \cite{lewis-etal-2020-bart}\footnote{We have considered the SOTA abstractive summarization LM, PEGASUS \cite{zhang2019pegasus}. Yet, the GPU memory consumption is too high for us to test it.}. The template generation problem much resembles summarization, except that generated \textit{template sequences} contain implicit structures. With the various denoising pre-training objectives, we believe that \textsc{BART} can capture the implicit structure within \textit{template sequences}, effectively model the dependencies among predicted entities, and produce rich semantics to reason over between slot names and entities. % \sam{probably worth more explicitly commenting on BART allowing semantic meaning from pre-training to be applied to entity extraction and slot filling} \textit{template sequences}.

\paragraph{Cross-attention guided copy mechanism}
To enhance \textsc{BART}'s capabilities to identify salient mentions in the input documents, we incorporate a copy mechanism based on cross-attention. As cross-attentions often imply saliency of input tokens, a naive approach of computing copy distributions $P_{\text{copy}}$ at time step $t$ over the input tokens is taking the mean of the last decoder layer's cross-attention across all heads, as mentioned in \citet{xu-etal-2020-self},

{
\vspace{-4mm}
\small
\begin{align}
     \alpha_{t, h} &=  \text{softmax}(\frac{(W_ss_t)^TW_ee}{\sqrt{d_k}}) \\
     P_{copy} &= \frac{\sum_{h} \alpha_{t, h} }{|H|},
\end{align}
}where $\alpha_{t, h}$ is the attention scores over input tokens at decoding step $t$ for head $h$. $W_s$ and $W_e$ are the projection matrices for the encoder and the decoder. $s_t$ is the decoder hidden states at step $t$, and $e$ denotes the encoder hidden states. 

However, recent studies have shown that attention heads are not equally important, and that some heads can be pruned out with a marginal decrease in overall performance \cite{voita-etal-2019-analyzing, NEURIPS2019_2c601ad9}. We hypothesize that the attention probabilities produced by insignificant attention heads may be noisy. Thus, computing copy distributions without these heads could improve the model's ability to infer the importance of each token in the input document. Motivated by this hypothesis, we propose \topkcopy\ , a copy mechanism where only the Top-$k$ important attention heads are used for computing copy distributions. Consider the formulation of multi-head attention, following the notation from \citet{ashish-etal-2017-attention}: %\sam{fix vertical margin spacing}

{
\vspace{-4mm}
\small
\begin{align}
    \text{MultiHead}(Q, K, V ) = \text{Concat}(\text{head}_1, ..., \text{head}_h)W^O& \\
    \text{head}_i = \text{Attention}(QW^Q_i, KW^K_i, VW^V_i).&
\end{align}
}
$W^Q_i, W^K_i, W^V_i \in \mathbb{R}^{d_{\text{model}} \times d}$ are the projection matrices for computing attention. $W^O \in \mathbb{R}^{hd_v \times d_\text{model}}$ is the matrix that allows interaction between different attention heads, where $h$ is the number of heads. To determine the importance of each attention head, we first transform $W^O$ to dimension $h \times d_v \times d_\text{model}$ (\Cref{eq:transform}), and then sum over the last two dimensions of $W^O$ (\Cref{eq:compute_score}), 

{
\small
\begin{align}
    W^O \in \mathbb{R}^{hd_v \times d_\text{model}} &\rightarrow 
    W^O \in \mathbb{R}^{h \times d_v \times d_\text{model}} \label{eq:transform} \\ 
    \text{score}_i &= \sum_{j, k} | W^O_{i, j, k} | \label{eq:compute_score}.
\end{align}
}
where $\text{score}_i$ denotes the significance score for head $i$. We take the attention heads with Top-$k$ highest significance scores in the last cross-attention layer, and use the mean of the attention probabilities outputted by these heads as the copy distribution as shown in equations~\ref{eq:topk} and \ref{eq:topk-att},

{
\vspace{-4mm}
\small
\begin{align}
    K &= \text{Top-}k(score) \label{eq:topk}\\
    P_{copy} &= \frac{\sum_{h \in K} \alpha_{t, h} }{k} \label{eq:topk-att}.
\end{align}
}
\paragraph{Objective function.}
The final probability $P_{\text{final}}$ of a word $w_t$ is a weighted sum of vocabulary distribution computed by \textsc{BART} $P_{\text{vocab}}$ and copy distribution $P_{\text{copy}}$,

{
\vspace{-4mm}
\small
\begin{align}
     P_{\text{final}}(w_t) = p_{\text{gen}}P_{\text{vocab}}(w_t) + (1 - p_{\text{gen}})P_{\text{copy}}(w_t).
\end{align}
}where $p_{\text{gen}} \in [0, 1]$ is the generation probability computed by passing the dot product of the mean encoder hidden state $\overline{e} = \frac{\sum^{n}_{i=0}e_i}{n}$%\violet{since you only encode everything once, there shouldn't be a notion of encoder state at time step $t$, no? Also, encoder should have k positions each corresponds to a word on the input side, which one you take as $e_t$? that last token?} 
and decoder hidden state $s_t$ at time step $t$ through the sigmoid function $\sigma$,

{
\vspace{-4mm}
\small
\begin{align}
     p_{\text{gen}} = \sigma(\overline{e} \cdot s_t)
\end{align}
}
Using the final probability distribution $P_{\text{final}}$, we can then compute the loss function as the average negative log likelihood of the target word $y_t$ over all timesteps, following \citet{see-etal-2017-get},

{
\vspace{-4mm}
\small
\begin{align}
     \mathcal{L} = \frac{1}{T}\sum_{t=0}^{T}-\log P_{\text{final}}(y_t).
\end{align}
}

\section{Experimental Setup}
\begin{table*}[t]
    \small
    \centering
    {
    \begin{tabular}{lccccccccc}
        \toprule
        
        & \multicolumn{3}{c}{\textbf{REE}} & 
        \multicolumn{3}{c}{\textbf{Binary RE}} & 
        \multicolumn{3}{c}{\textbf{4-ary RE}} \\ \midrule
        Model      & Prec. & Rec. &  F1 &  Prec. & Rec. &  F1 & Prec. & Rec. &  F1    \\ 
        \midrule
        NST \cite{du-cardie-2020-document}  & 
        56.82 & 48.92 & 52.58 & - & - & - & - & - & -\\
        TANL \cite{paolini2021structured} & 64.89 &  47.75 &  55.02 &  0.74 &  0.67 &  0.62 &  0.00 &  0.00 &  0.00\\
        GRIT    \cite{du-2020-grit} &
        64.19 & 47.36 & 54.50 & - & - & - & - & - & -\\
        \dygiepp~ \cite{wadden-etal-2019-entity} &
        57.04 & 46.77 & 51.40  & 2.9 & 12.8 & 3.8 & - & - & - \\
        \scirexpipeline{} \cite{jain-etal-2020-scirex} & -
         & - & - & 6.5 & \textbf{41.1} & 9.6 & 0.7 & \textbf{17.3} & 0.8\\ 
        
        \cmidrule(lr){1-10} 
        
        \modelshort~&  \textbf{68.55} & \textbf{49.90} & \textbf{57.76} & \textbf{17.11} & 13.56 & \textbf{14.47}$^{*}$ & \textbf{3.19} & 4.26 & \textbf{3.55}$^{*}$ \\

        \bottomrule
    \end{tabular}
    }
    \vspace{-2mm}
    \caption{Performance comparison on role-filler entity extraction, binary and 4-ary relation extraction tasks. TANL results are re-implemented and evaluated by ourselves. \modelshort~ outperforms all previous systems on REE, binary RE, and 4-ary RE. Statistical significance over previous best systems computed using the paired bootstrap procedure \cite{berg-kirkpatrick-etal-2012-empirical} is indicated with $^* (\textit{p}<.01)$.}%, $^{\dagger} (\textit{p}=.002)$, and $^{\ddagger} (\textit{p}=.007)$.}
    \label{tab:main}
    \vspace{-5mm}
\end{table*}
\subsection{Dataset and Evaluation Metric}
Experiments are conducted on two English datasets: MUC-4 (\citeyear{muc-1992-message}) for the role-filler entity extraction task and SciREX \cite{jain-etal-2020-scirex} for the binary and 4-ary end-to-end relation extraction tasks. MUC-4 contains 1700 documents, with on average about 400 tokens per document. Documents are annotated with zero to multiple event templates. As per \citet{du-2020-grit}'s pre-processing, we have a 13:2:2 split on the documents for train, development, and test, respectively. We evaluate the REE task on this dataset using the entity-level metric, CEAF-REE \cite{du-2020-grit}. The metric aligns predicted entities with gold entities using Kuhn–Munkres algorithm \cite{kuhn1955hungarian, munkres1957algorithms}, where a predicted entity is considered correct if and only if its corresponding mentions are a subset of the aligned gold entity's mentions.

The \textsc{SciREX} dataset\footnote{https://github.com/allenai/SciREX} consists of scientific articles, with entity, coreference, and relation annotations. With an average token count of about 5700, the articles are significantly longer than the documents in MUC-4. We use the pre-processed data from \citet{jain-etal-2020-scirex}, which contains 306 documents for training, 66 for validation, and 66 for testing. In contrast to conventional relation extraction datasets, such as ACE05, relations are not typed in \textsc{SciREX}.
Hence, the official \textsc{SciREX} evaluator \cite{jain-etal-2020-scirex} only considers the correctness of predicted entities and entity types\footnote{There are 4 entity types: \textsc{Material}, \textsc{Metric}, \textsc{Task}, and \textsc{Method}.} in each relation. Predicted entities are aligned with gold entities based on mention overlap. When the entities are aligned, predicted relations are aligned with gold relations accordingly. A predicted relation is correct if and only if both the associated entities and the entity types match the aligned gold relation.
%\sam{By nature of this section, it's pretty difficult to transition between sentences, but would flow much better if you do find transitions}
\subsection{Baselines}
We compare our method with the following competitive baseline systems. 

\textbf{NST} \cite{du-cardie-2020-document} builds multi-granularity representations on documents, and utilizes gate mechanism to fuse representations of different granularity. 

\textbf{TANL} \cite{paolini2021structured} augments sequential labels with input sentences, allowing it to be applied to various structured prediction tasks\footnote{Since the source code of TANL has not been released by the time we conducted experiments, we re-implemented it by closely following the method described in \citet{paolini2021structured}.}. 

\textbf{GRIT} \cite{du-2020-grit} shares transformer parameters between the encoder and the pointer network decoder, and is the SOTA system for the REE task on the MUC dataset. 

\textbf{\dygiepp}\ \cite{wadden-etal-2019-entity} is a span-based multi-task IE framework jointly trained on relation extraction, named entity recognition, and coreference resolution. 

\textbf{\scirexpipeline} \cite{jain-etal-2020-scirex} is the SOTA framework for end-to-end binary and 4-ary relation extraction on \textsc{SciREX}. The pipeline is composed of 4 components: mention identification, mention clustering, salient entity cluster identification, and relation classification.  

In terms of the pre-trained language models used, \textsc{BERT-base} \cite{Devlin_2019} is used for \textbf{NST}, \textbf{\dygiepp}, and \textbf{GRIT}. \textbf{\scirexpipeline} is fine-tuned on \textsc{SciBERT} \cite{beltagy-etal-2019-scibert}. We replace \textsc{T5} \cite{JMLR:v21:20-074} with \textsc{BART-base} for \textbf{TANL} for a fair comparison with our method.

\subsection{Implementation details}
The proposed models are optimized using AdamW \cite{loshchilov2018decoupled} with learning rate 5e-5 and weight decay 1e-5. We used grid search to find the best $k$ for \topkcopy \ \ and found that $k=10$ yields the best overall performance across REE and RE. The maximum input sequence length for RE and REE are 1024 and 512, respectively. During inference time, all generative models used beam search with a beam width of 4.

% For the re-implemented TANL, we optimize it with a linear learning rate of 5e-4, following \citet{paolini2021structured}. The only change we made is replacing AdamW with AdaFactor, a more suitable optimizer for \textsc{T5}, as suggested in \citet{JMLR:v21:20-074}. 

\section{Results and Analysis}

\subsection{Main Results} %\violet{the RE results are miserably low... did Jain et al. really published their paper with such low numbers? Or did you rerun their code and got different numbers?} \ssteeve{Yes, Jain et al. published these numbers. SciREX is a pretty challenging dataset.}

\label{sec:main_results}
% \Steeve{Maybe we can show significance test somehow?}
\Cref{tab:main} summarizes the main results on role-filler entity extraction, binary, and 4-ary relation extraction. \modelshort~ establishes new state-of-the-art scores on all three tasks, outperforming the previous best models by an absolute F1 of 3.26\%, 4.8\%, and  2.7\%. The improvements demonstrate the effectiveness of our approach in formulating document-level EE tasks into template generation tasks. Although \modelshort~ scores the highest F1 across all three tasks, \scirexpipeline~ does achieve the highest recall on both RE tasks. This can be explained by the fact that our model can only encode the first 1024 sub-tokens of each \textsc{SciREX} document, which is merely 17\% of the average sub-token count per document. This makes it challenging for \modelshort~ to identify relations that lie in the latter 83\% of each document. In the future, we can extend \textsc{BART}'s positional embedding matrix to enable \modelshort~ to encode longer documents. Additionally, we set the maximum input sequence length to 512 for \modelshort~ for fairer comparisons with \scirexpipeline~. We obtain F1 scores of 11.94\%
and 2.18\% on binary and 4-ary relation extraction, respectively. This confirms the advantage of our model on the relation extraction tasks.

While TANL performs worse than our model on REE, it is still able to achieve a higher score than GRIT. This suggests that augmenting decoding targets with label names provides useful semantics, whereas adding input documents to decoding targets may not yield better results in the REE task. We also observe that TANL scores extremely low on both RE tasks, where 58\% of the binary relations and 26 \% of the 4-ary relations in the decoding targets are filtered out due to exceeding maximum sequence length of \textsc{BART}. Out of the remaining relations, 57\% of the binary relations and 78\% of the 4-ary relations have at least one entity removed in the decoding targets due to its long distance from the first-appearing entity\footnote{Please refer to \Cref{sec:TANL_targets} for more details.}, suggesting that TANL's poor performance on RE tasks is due to scarcity of gold labels. This reflects that TANL is ill-suited for document-level EE tasks.
% \begin{table*}[h!]
%     \small
%     \centering
%     {
%     \begin{tabular}{lccccccccc}
%         \toprule
        
%         & \multicolumn{3}{c}{\textbf{REE}} & 
%         \multicolumn{3}{c}{\textbf{Binary RE}} & 
%         \multicolumn{3}{c}{\textbf{4-ary RE}} \\ \midrule
%         Model      & Prec. & Rec. &  F1 &  Prec. & Rec. &  F1 & Prec. & Rec. &  F1    \\ 
%         \midrule
%         \modelshort~&  \textbf{68.55} & 49.90 & \textbf{57.76} & \textbf{17.11} & \textbf{13.56} & \textbf{14.47} & 3.19 & \textbf{4.26} & \textbf{3.55} \\
%         % \cmidrule(lr){1-10}
%         ~ \textsc{TopK copy} $\rightarrow$ \textsc{Naive Copy} & 64.65 & 50.10 & 56.45 & 12.35 & 10.98 & 11.22 & 2.13 & 1.06 & 1.42 \\
%         ~ \textsc{TopK copy} $\rightarrow$ SAGCopy & 61.19 & 44.42 & 54.47 & 12.20 & 10.83 & 11.17 & 1.06 & 2.13 & 1.41  \\
%         % \cmidrule(lr){1-10}
%         ~ w/o \topkcopy\ & 55.17 & \textbf{56.36} & 55.76 & 16.07 & 11.38 & 12.63 & \textbf{4.25} & 2.70 & 3.00 \\
%         ~ numeric slot name & 62.74 & 51.08 & 56.31& 10.71 & 7.37 & 8.22 & 2.13 & 0.53 & 0.85 \\
%         % ~ no slot name &  \\
%         % ~ \textsc{BART} $\rightarrow$ \textsc{T5} &  \\
%         % ~ \textsc{BART} $\rightarrow$ \textsc{RoBERTa} + transformer  & \\

%         \bottomrule
%     \end{tabular}
%     }
%     \vspace{-2mm}
%     \caption{Ablation study on removing and replacing different components of \modelshort~.}
%     \label{tab:ablation}
%     \vspace{-5mm}
% \end{table*}
\begin{table}[t!]
    \small
    \centering
    {
    \begin{tabular}{lccc}
        \toprule
        
        % & \multicolumn{3}{c}{\textbf{REE}} & 
        % \multicolumn{3}{c}{\textbf{Binary RE}} & 
        % \multicolumn{3}{c}{\textbf{4-ary RE}} \\ \midrule
        Model       &  REE &  Binary RE & 4-ary RE \\
        \midrule
        \modelshort~   & \textbf{57.76} & \textbf{14.47}  & \textbf{3.55} \\
        % \cmidrule(lr){1-10}
        ~ $\rightarrow$ \textsc{Naive Copy}  & 56.45 & 11.22 & 1.42 \\
        ~ $\rightarrow$ SAGCopy & 54.47 & 11.17 & 1.41  \\
        % \cmidrule(lr){1-10}
        ~ w/o \topkcopy\ & 55.76 & 12.63 & 3.00 \\
        ~ numeric slot name & 56.31 & 8.22 & 0.85 \\
        % ~ no slot name &  \\
        % ~ \textsc{BART} $\rightarrow$ \textsc{T5} &  \\
        % ~ \textsc{BART} $\rightarrow$ \textsc{RoBERTa} + transformer  & \\

        \bottomrule
    \end{tabular}
    }
    \vspace{-2mm}
    \caption{Ablation study on removing and replacing different components of \modelshort~.}
    \label{tab:ablation}
    \vspace{-5mm}
\end{table}

We observe extremely low performances across all systems on both tasks of \textsc{SciREX}, even though \modelshort~ outperforms the baseline systems significantly. This is mainly caused by the characteristics of the \textsc{SciREX} dataset. First, syntactic characteristics specific to scientific journals, such as algorithm blocks, result in the unusually long sequences in the \textsc{SciREX} dataset despite best parsing efforts. 
Additionally, another feature frequently seen in scientific journals is the use of table and figure captions. Since captions are not included as part of the input text, the number of accepted relations decreases drastically.

\begin{figure}[t]
    \centering
    \begin{subfigure}{0.48\textwidth}

    \includegraphics[width=.95\linewidth]{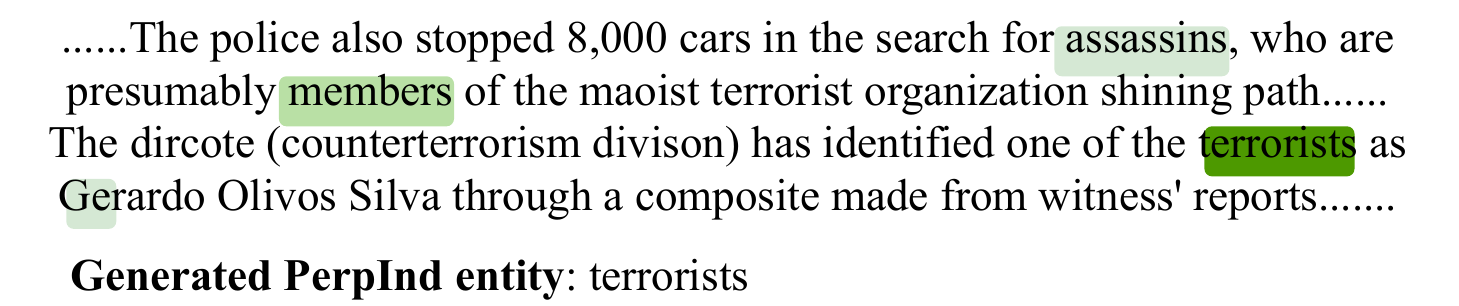}
    \caption{Copy distribution produced by \textsc{Naive Copy}. }
    \end{subfigure}
    \par\bigskip
    \begin{subfigure}[b]{0.48\textwidth}
    \includegraphics[width=.95\linewidth]{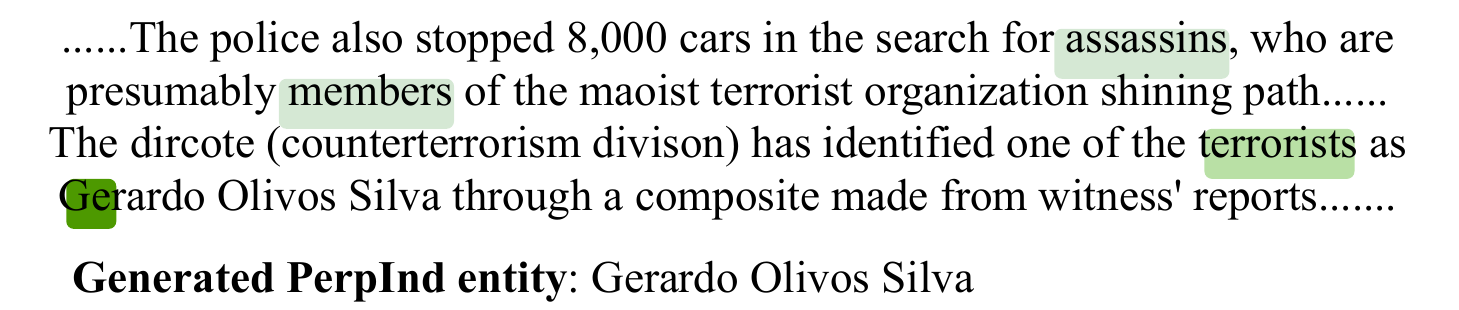}
    \caption{Copy distribution produced by \topkcopy~.}
    \end{subfigure}
    \par\bigskip
    \begin{subfigure}[b]{0.48\textwidth}
    \begin{center}
    \includegraphics[width=.5\linewidth]{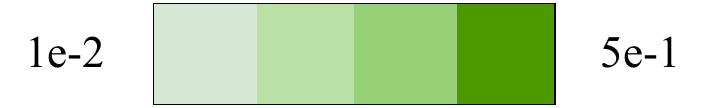}
    \caption{The darker the color, the higher the probability.}
    \end{center}
    \end{subfigure}
    
    \caption{\topkcopy~ produces a more reliable copy distribution $P_{copy}$ than that computed by \textsc{Naive Copy} in a MUC-4 example. Given an input document and decoded tokens ``<s>\MyColorBox[yellow!10]{<SOT>}\MyColorBox[cyan!10]{<SOSN>}\textsc{PerpInd}\MyColorBox[cyan!10]{<EOSN>}\MyColorBox[red!10]{<SOE>}'', the gold \textsc{PerpInd} entity is ``Gerado Olivos Silva''. However, ``terrorist'' is assigned the highest copy probability computed by \textsc{Naive Copy}, leading to incorrect entity extracted. %By only considering the attentions computed by the top-$k$ most significant cross-attention heads
    Conversely, \topkcopy~ assigns the highest $P_{copy}$ to the head token of the gold entity, ``Ger'', resulting in successful extraction of the correct entity eventually. }
    \label{fig:attention_comparison}
    \vspace{-5mm}
\end{figure}

\subsection{Performance Analysis}

% \paragraph{The capacity of BART}
% \ \Steeve{Need to run experiments to compare with randomly initialize BART and maybe BERT+transformer.}\
\begin{figure*}[t]
    \centering
    \begin{subfigure}[t]{0.31\textwidth}

    \includegraphics[width=.95\linewidth]{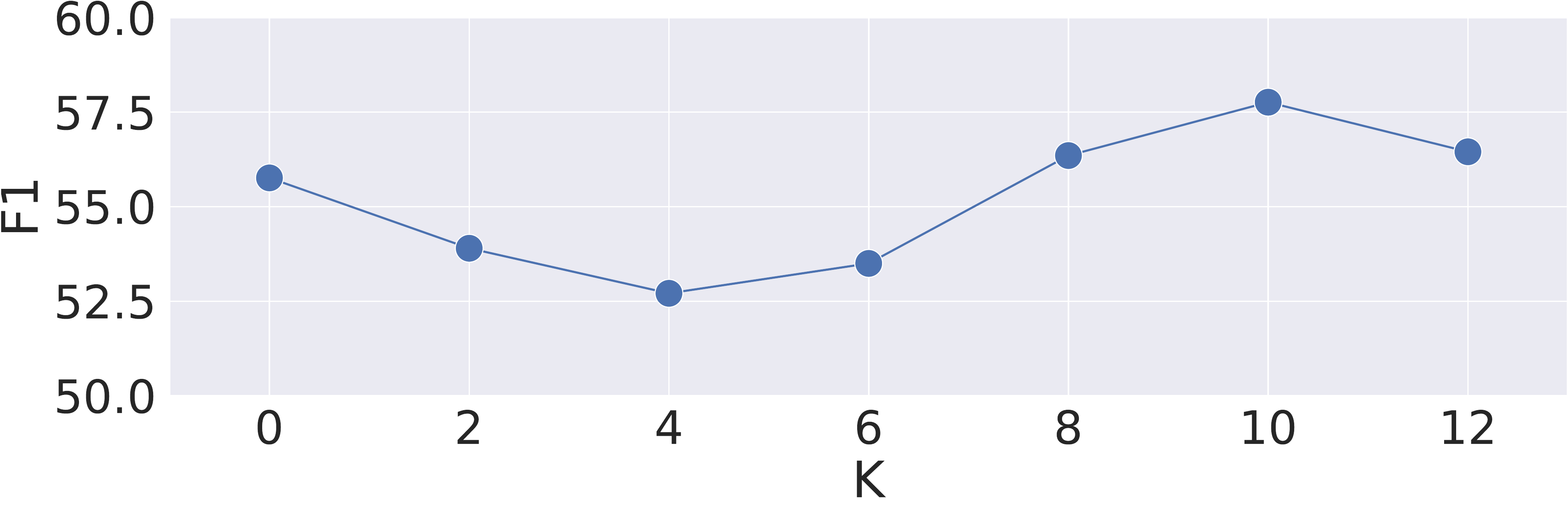}
    \caption{Role-filler entity extraction.}
    \end{subfigure}
    ~
    \begin{subfigure}[t]{0.31\textwidth}
    \includegraphics[width=.95\linewidth]{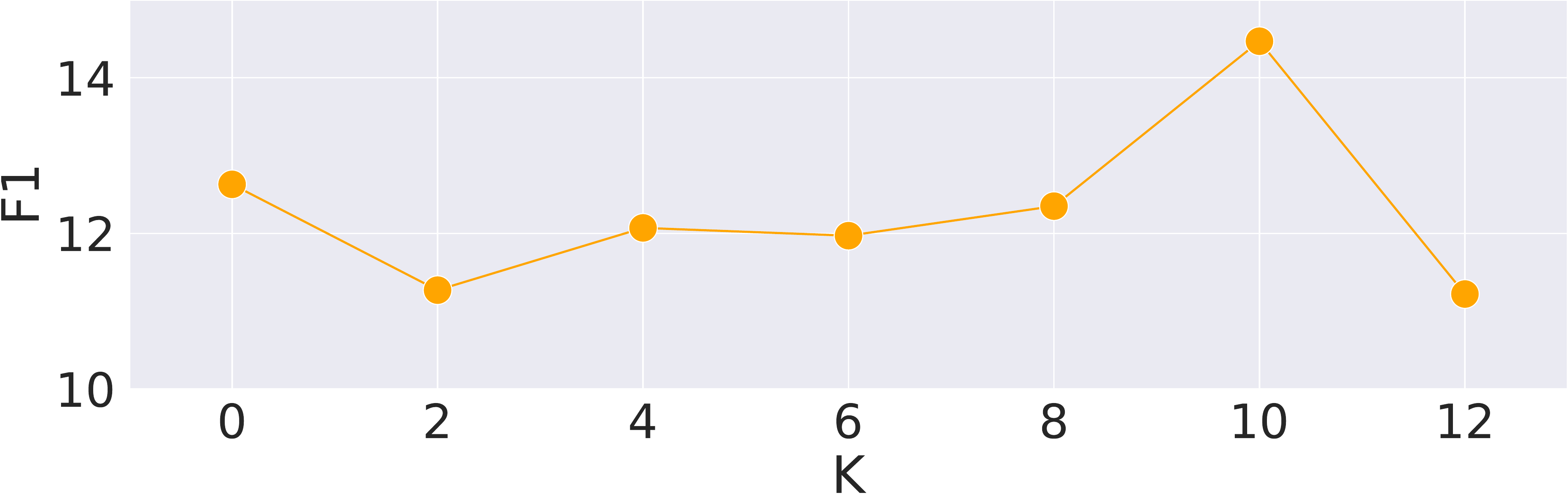}
    \caption{Binary relation extraction.}
    \end{subfigure}
    ~
    \begin{subfigure}[t]{0.31\textwidth}
    \includegraphics[width=.95\linewidth]{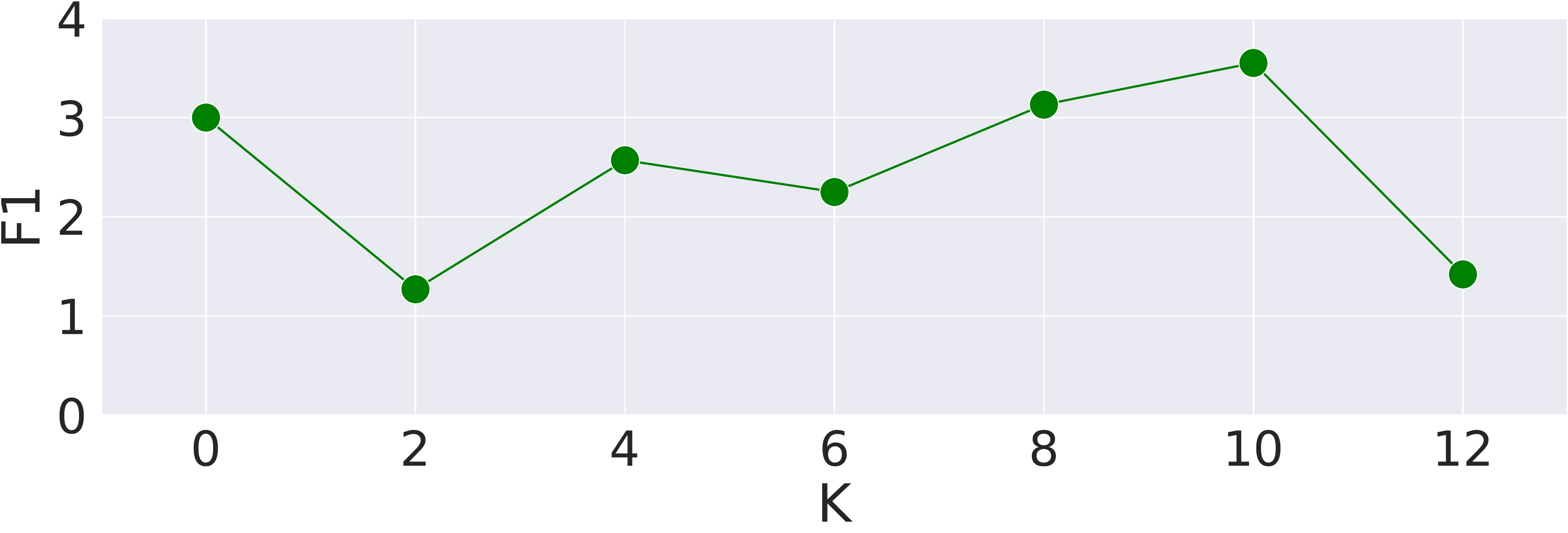}
    \caption{4-ary relation extraction.}
    \end{subfigure}
    
    \caption{Effect of $K$. We compare the test set F1 score on MUC-4 REE, \textsc{SciREX} binary and 4-ary RE tasks with regard to different $K$. $K = 0$ is equivalent to removing \topkcopy~.}
    \label{fig:effect_of_k}
    \vspace{-5mm}
\end{figure*}

\paragraph{Ablation Study} %\violet{table 2 might show better if we only show F1 score.}
We conducted ablation studies by replacing the \topkcopy~ module with other copy mechanisms. \textbf{\textsc{Naive Copy}} refers to computing copy distributions with the attentions from all cross-attention heads. \textbf{SAGCopy} \cite{xu-etal-2020-self} utilizes encoder self-attention to compute centrality scores for measuring the saliency of each input token. As shown in \Cref{tab:ablation}, we found that \textsc{Naive Copy} leads to performance drop on all three tasks, especially on binary and 4-ary relation extractions. \textbf{\textsc{Naive Copy}} achieving scores even lower than fine-tuning \textsc{BART} alone (i.e. w/o \topkcopy~ ) reflects that copy mechanisms may mislead models to copy incorrect input tokens. A qualitative example of the difference between \topkcopy~ and \textsc{Naive Copy} demonstrated in \Cref{fig:attention_comparison} validates our hypothesis. Quantitatively, examining MUC-4 test set predictions, there are 79 cases where \topkcopy~ corrects the misguidance of Naive Copy, while only 32 cases where new errors are introduced by \topkcopy~. %\violet{do we have any quantitative results?} 
For both REE and RE, adding SAGCopy leads to performance drop, suggesting that the centrality scores of input tokens may not be an ideal feature for these tasks.

We also experimented with replacing the original slot names with numeric slot names (i.e. converting \textsc{PerpInd} to \textsc{<ROLE\_1>}, \textsc{PerpOrg} to \textsc{<ROLE\_2>}, and etc). This conversion removes the semantics of slot names in the decoding targets. While little performance drop was observed on the REE task, using numeric slot names resulted in the worst performance on binary and 4-ary relation extraction tasks, which could be a result of strong slot dependencies in RE in comparison with REE. In RE, slots are directly semantically related to other slots in each template whereas slots in REE are relatively independent. This shows that slot name semantics are useful for template generation tasks with strong slot dependencies in each template. 
Finally, we conducted ablation studies on different variations of templates as decoding targets. Specifically, three variations are tested on the REE task: (1) We merge entities of the same role names into the same ``slot''. (e.g. transforming the decoding targets from ``\MyColorBox[cyan!10]{<SOSN>}PerpInd\MyColorBox[cyan!10]{<EOSN>}\MyColorBox[red!10]{<SOE>}Alice\MyColorBox[red!10]{<EOE>} \MyColorBox[cyan!10]{<SOSN>}PerpInd\MyColorBox[cyan!10]{<EOSN>}\MyColorBox[red!10]{<SOE>}Bob\MyColorBox[red!10]{<EOE>}'' to ``\MyColorBox[cyan!10]{<SOSN>}PerpInd\MyColorBox[cyan!10]{<EOSN>}\MyColorBox[red!10]{<SOE>}Alice; Bob \MyColorBox[red!10]{<EOE>}''). (2) Based on (1), all slot names, such as ``PerpInd'' and ``PerpOrg'', are replaced with the same special token ``<ROLE>''. (3) We use the same decoding targets as GRIT's. These three settings achieve test set F1 scores of 56.65, 54.16, and 52.55, respectively. The results suggest that differentiating entities with different entity types helps improve the performance. Furthermore, comparing with the results in \Cref{tab:main}, we found that GRIT performs better than our system, reflecting that a pointer network-based model, which has with smaller search space than ours, is more advantageous when using the same decoding targets.

\paragraph{Impact of the Amount of Training Data}
\begin{figure}[t]
    \centering
    \includegraphics[width=.95\linewidth]{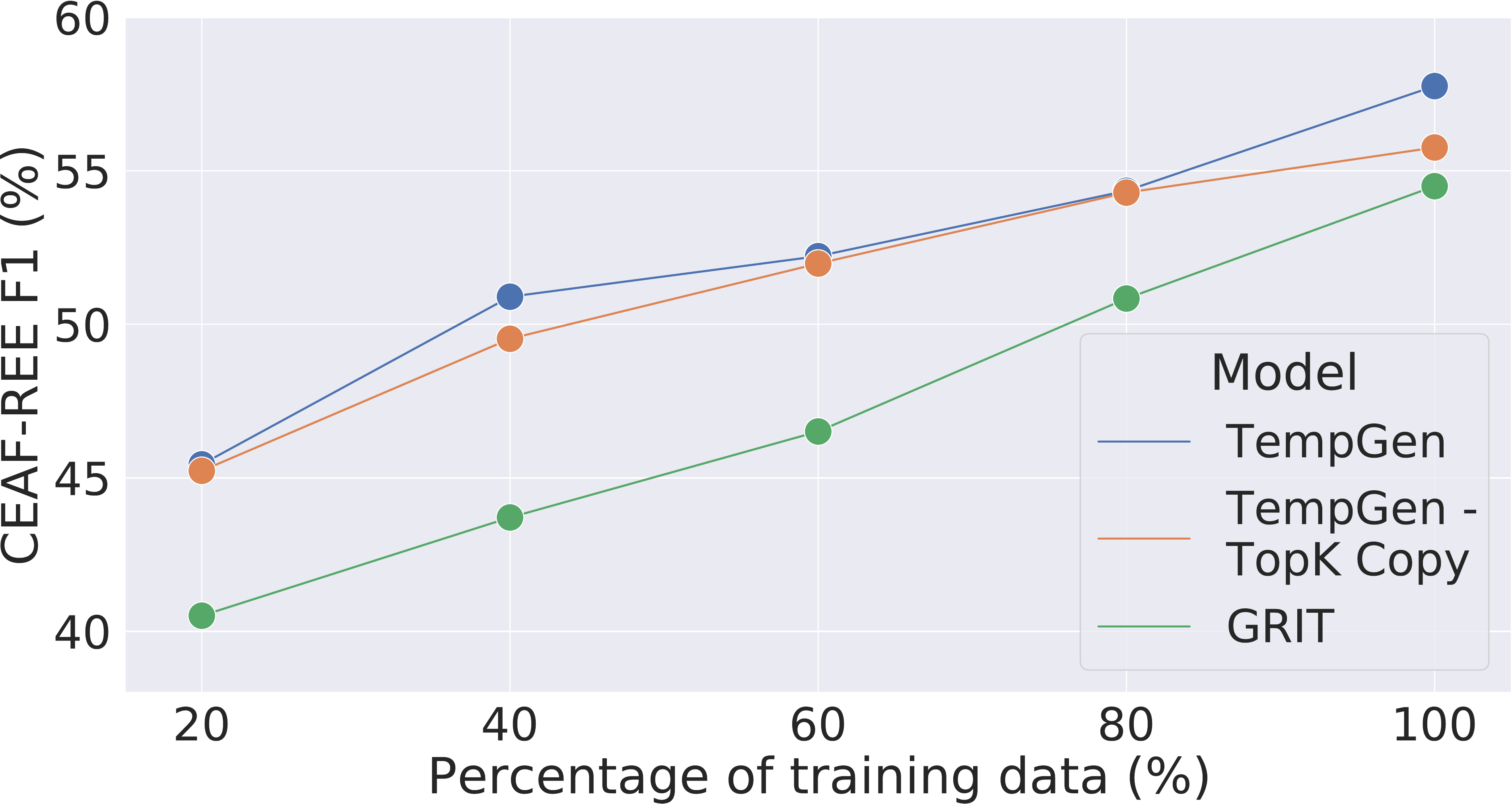}
    \caption{REE test set performance on MUC-4 with regard to different amount of training data.}
    \label{fig:data_efficiency}
\end{figure}
To test the data efficiency of our approach, we compared \modelshort~ and \modelshort~ - \topkcopy~ with GRIT on the REE task using different amount of MUC training data. As seen in \Cref{fig:data_efficiency}, both \modelshort~ and \modelshort~ - \topkcopy~ outperform GRIT across all settings with a slightly larger performance margin in low resource settings. This indicates that our approach is more data-efficient compared to the previous SOTA system on REE.

\paragraph{Impact of $K$ Cross-attention Heads}
\Cref{fig:effect_of_k} shows our model's change in performance conditioned on various values of $K$ in the \topkcopy~ mechanism. Consistent with our results in \Cref{sec:main_results}, we see that removing some of the cross-attention heads (12 $\rightarrow$ 10) can lead to performance gain due to the filtered noise brought by unimportant attention heads. However, we noticed a drop in performance across all three tasks for lower values of $K$, suggesting that beneficial cross-attention heads are removed. Interestingly, performance drops immediately as $K$ decreases below 10, suggesting that only a small portion of the cross-attention heads are unimportant. The trend is consistent with \citet{NEURIPS2019_2c601ad9}'s results where
pruning cross-attention heads to a certain extent can easily result in performance drop. Additionally, the model with no copy mechanism ($K=0$) outperforms the model with few attention heads ($K \in \{2,4,6\}$), suggesting that the copy distributions obtained from not sufficiently informative cross-attentions can mislead the model.

\subsection{Qualitative Analysis}
%Comparing outputs from \modelshort~ with other baseline systems, 
The following qualitative analysis provides intuition for our model's ability to capture dependencies across entities and utilize slot name semantics. %\sam{something like "better extract important label semantics and catpure dependencies across entities"} 

\paragraph{Cross-entity Dependencies} To validate our approach's capability to capture cross-entity dependencies, we considered binary relations on \textsc{SciREX} where at least one of the associated entities is involved in multiple relations. The dependencies among entities are better captured by the model that predicts fewer unlikely relations. %\sam{Not sure I understood this fully} 
 Comparing the test set outputs of \modelshort~ and \scirexpipeline~, we see that 13131 errors made by \scirexpipeline~ are corrected by our model, which only introduces 604 errors. This result demonstrates the strength of \modelshort~ in modeling cross-entity dependencies.% and avoiding extracting unlikely relations.

\begin{figure}[t]
    \centering
    \includegraphics[width=.95\linewidth]{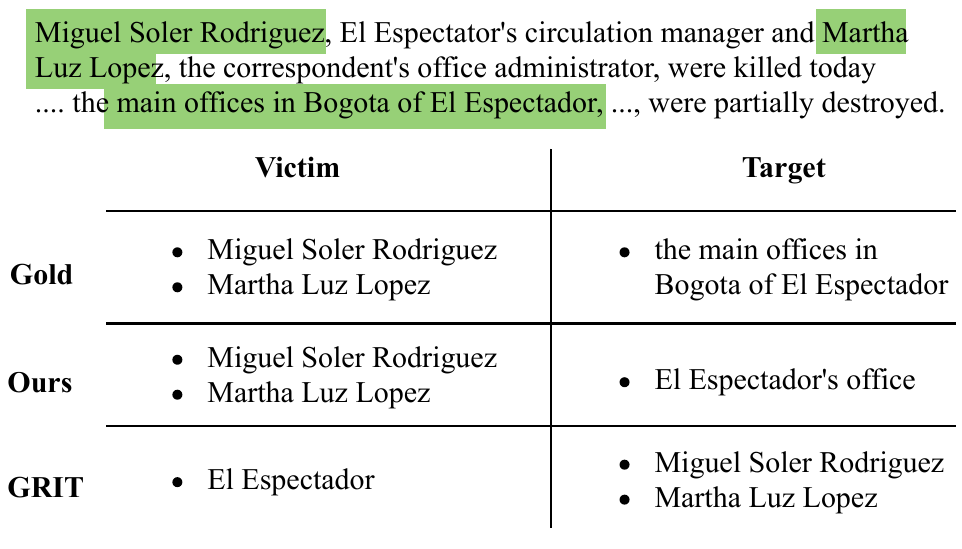}
    \caption{An example showing how GRIT  misidentifies the \textsc{Victim} entities and \textsc{Target} entities, likely due to the lack of role type semantics. Here, \textsc{Victim} entities are the people attacked, and \textsc{Target} entities are the objects compromised.}
    \vspace{-4mm}
    \label{fig:ree_qualitative}
\end{figure}

\paragraph{Importance of Label Semantics} Comparing the test set predictions between \modelshort~ and GRIT on the MUC-4 REE task, we see that our approach better distinguishes confusing entities such as \textsc{Victim} and \textsc{Target} entities. %\sam{not sure if I changed the meaning here, if so feel free to just reject} 
As shown in the example in \Cref{fig:ree_qualitative}, GRIT incorrectly predicts the two victims, ``Miguel Soler Rodrigues'' and ``Martha Luz Lopez'', as \textsc{Target} entities. It also misidentifies ``El Espectador'', a newspaper company, as a victim of the attack. In contrast, \modelshort~ is able to recognize the roles of the two victims. Even though it's not an exact match, the predicted \textsc{Target} entity had a correctly identified role type with similar semantic meaning compared to the gold label.

\section{Inference time comparison}
\begin{figure}[t]
    \centering
    \includegraphics[width=.95\linewidth]{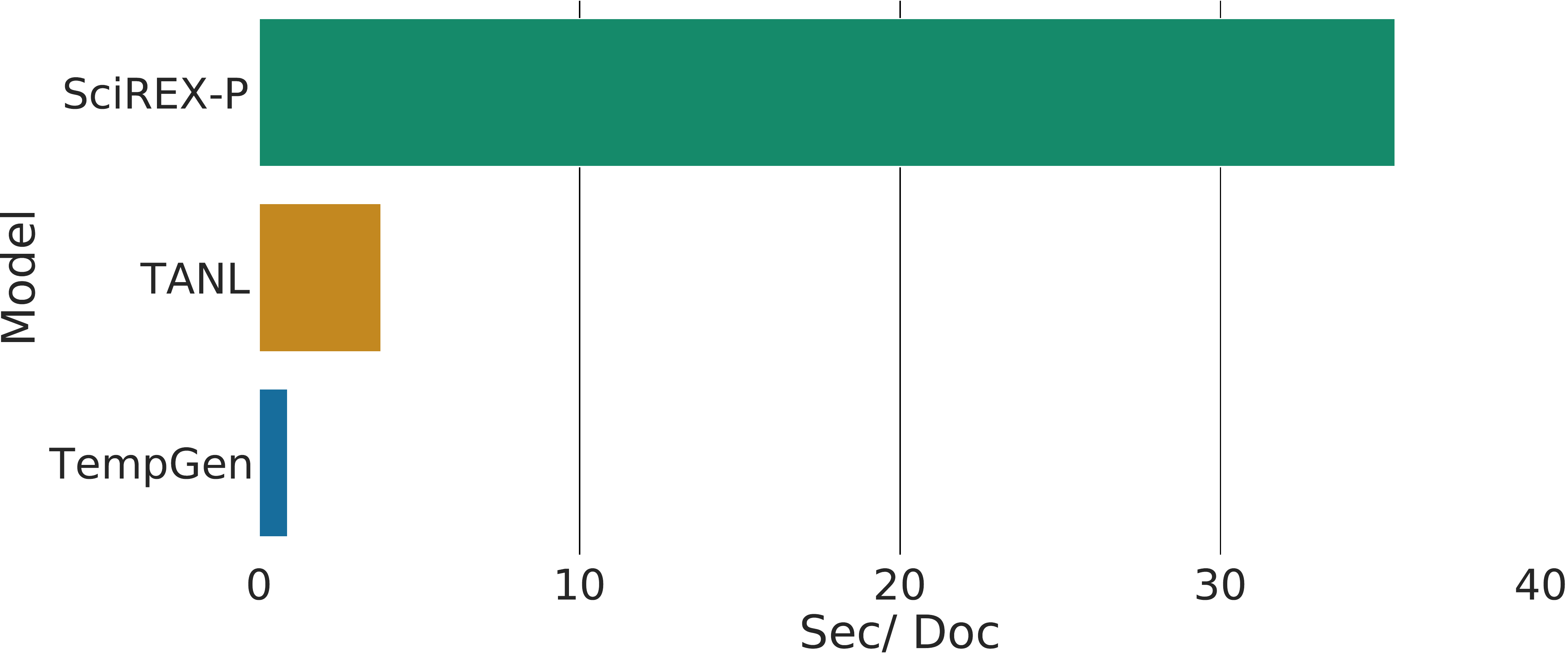}
    \caption{Inference time comparison on the \textsc{SciREX} 4-ary RE task.}
    \vspace{-4mm}
    \label{fig:inference_time}
\end{figure}

As discussed earlier, \modelshort~ can significantly reduce the exponential computational complexity of document-level N-ary relation identification. To illustrate this, we compared the inference time between \modelshort~ and two other systems, TANL and \scirexpipeline~, on the \textsc{SciREX} 4-ary RE task. As shown in \Cref{fig:inference_time}, \modelshort~ drastically shortens the inference time by around 39 times compared to \scirexpipeline~. TANL also runs much faster than \scirexpipeline~, but is still around 4 times slower than \modelshort~. This is resulted from the fact that TANL generates the entire input document in addition to entity and relation labels, which is much longer than \modelshort~'s generated sequences.

\section{Number of Parameters}
\begin{figure}[h]
    \centering
    \includegraphics[width=.95\linewidth]{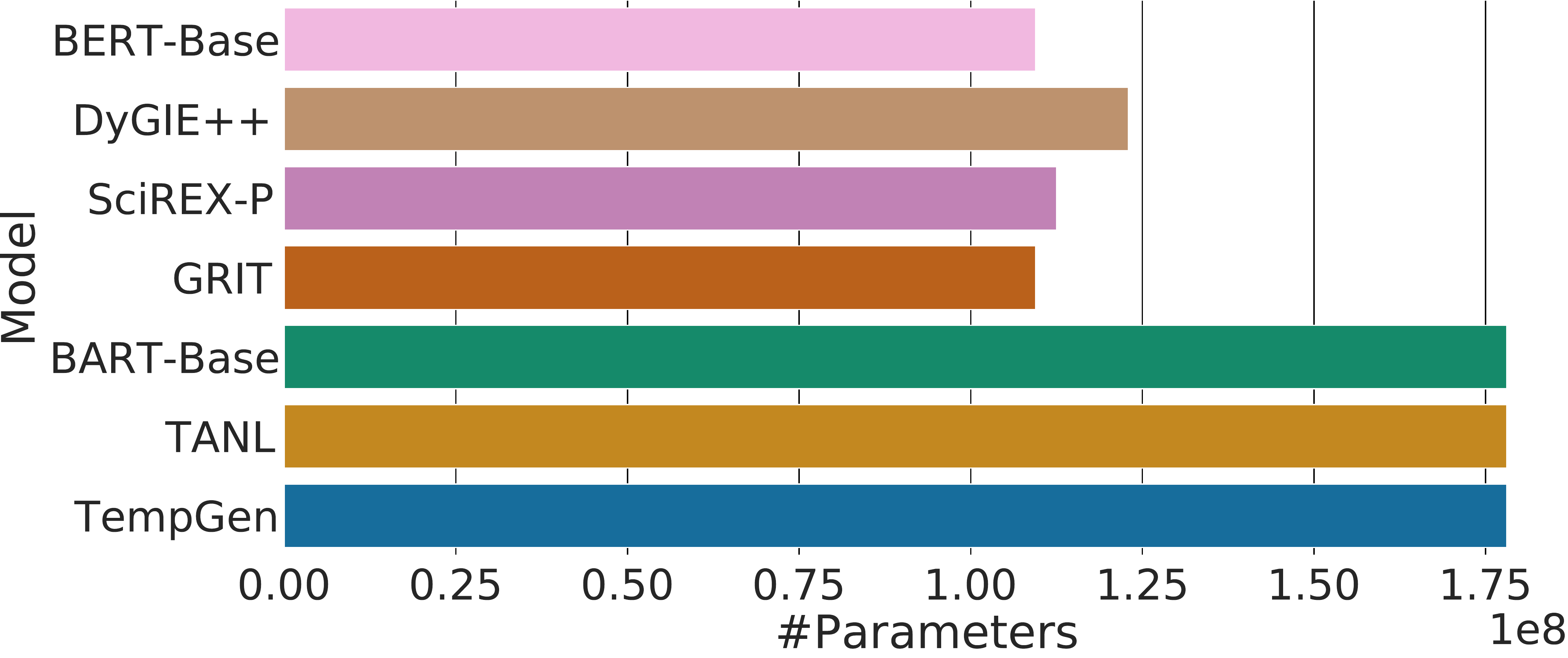}
    \caption{Number of parameters of different systems.}
    \vspace{-4mm}
    \label{fig:parameters}
\end{figure}

\Cref{fig:parameters} shows the number of parameters of different models. GRIT, with the same size of \textsc{BERT-base}, has the least number of parameters among all models. \dygiepp~ and \scirexpipeline~ have slightly more parameters than GRIT due to the additional linear layers for constructing classifiers. The two generative models, TANL and \modelshort~, have the most parameters, thanks to the larger vocab size (30522 $\rightarrow$ 50265), larger positional embedding matrix (512 $\rightarrow$ 1024), and cross-attention modules in \textsc{BART-base}.

% \sam{can we move figure 6 above this section ?}

%\sam{Just a general note, anything that we did before writing the paper, like experiments, make sure to use past tense "we conducted", "we considered", "we built". But anything regarding the results, or the model, or addressing the reader, it's present tense "the model performs.." "results show ..." "Our paper aims to..." } \ssteeve{Got it!}

% \subsection{Case studies}

% \paragraph{Natural language as decoding targets}

% \begin{figure}[h]
%     \centering
%     \includegraphics[width=.95\linewidth]{example-image-a}
%     \caption{A visualization of top-k cross-attention guidance v.s. naive cross-attention guidance}
%     \label{fig:attention}
% \end{figure}
\section{Related Works}
In the following sections, we will first discuss a few important works on the REE task and document-level RE task. Then, we will dive into a few works that uses a similar sequence generative approach to various document-level IE tasks.

\subsection{Role-filler Entity Extraction}
Document-level REE has been explored in recent works using a variety of model architectures. \citet{du2020document} formulates the task as a sequence tagging problem, and trains layered classifiers as sequence readers on multiple granularities. In contrast, GRIT \cite{du-2020-grit} formulates the problem as sequence generation, and employs a single transformer layer whose parameters are shared between encoder and decoder to enrich semantics in the shared parameters. A pointer selection network is used for the final layer of decoding. 

\subsection{Document-level Relation Extraction}
Due to long-term dependencies that often span over hundreds of tokens, capturing entity relations have proven to be a challenging task. One approach was constructing a document-level graph from sentence encoding, then extracting entity relations from edge representations in the graph \cite{christopoulou-etal-2019-connecting}. Other works such as \citet{jia-etal-2019-document} layer classifiers in a pipeline architecture to obtain hierarchical representation of $N$-ary relations.

\subsection{IE as Sequence Generation}
Recently, there has been an increasing number of works framing information extraction tasks as sequence generation problem. \citet{zeng-etal-2018-extracting} formed triple extraction as a sequence generation task and adopted a RNN-based model with copy mechanisms. To encourage the faithfullness of the extracted triplets, \citet{ye2020contrastive} designed a triplet contrastive training objective. These works focus on sentence-level triplet extraction, while our work extracts role-filler entities and entity relations at the document level. \citet{li2021documentlevel,hsu2021event} formulates the document-level event argument extraction task as a conditional generation problem by providing event ontology. However, their work cannot be applied to REE or RE due to the lack of ontology for role-filler entities and relations. \citet{du-2020-grit} relied on a pointer-network-based decoder \cite{10.5555/2969442.2969540} to extract event role-filler entities, and the parameters of \textsc{BERT} \cite{Devlin_2019} is shared between the encoder and the decoder. Nevertheless, their method cannot incorporate role labels, whereas our approach can take advantage of the label semantics.

\citet{paolini2021structured} uses a very similar generative approach, which constructs decoder targets by inserting text markers and labels around entity mentions in the input sentence. The key idea is that augmenting the decoder targets with original input sentence and labels provides stronger semantics to the model. Unfortunately, modeling cross-entity dependencies remains a challenge as entities are further apart in their decoding targets. We instead transform annotations into \textit{template sequences} as decoding targets, where distances between entities are significantly shortened. Thus, our approach alleviates the scalability challenge of capturing cross-entity dependencies at the scale of documents. Additionally, our approach differs in that the length of our decoder targets is significantly shorter, allowing the non-truncated decoder targets to fit in pre-trained language models. In contrast, for their method, the gold decoder targets are guaranteed to be longer than corresponding input document. Since the length of input tokens are often greater than the max sequence length of pre-trained language models for document-level EE, a great portion of the gold labels will be skipped using \citet{paolini2021structured}'s method. 
\section{Conclusion}
We have proposed \modelshort~, a framework that frames document-level REE and RE tasks as a template generation task. A copy mechanism that takes the top-$k$ important cross-attentions as copy distributions is incorporated into \textsc{BART} for capturing key information in the input document. Experimental results on MUC-4 and \textsc{SciREX} showed that \modelshort~ outperforms prior approaches on role-filler entity extraction and end-to-end document-level relation extraction tasks. Under different amount of training data, \modelshort~ demonstrates robustness across all settings, while being advantageous in lower-resource regime.

\section*{Acknowledgements}
We appreciate insightful feedback from PLUSLab members and the anonymous reviewers. This research was sponsored by the Intelligence Advanced Research Projects Activity (IARPA), via Contract No. 2019-19051600007. The views and conclusions of this paper are those of the authors and do not reflect the official policy or position of IARPA or the US government.

\bibliography{anthology,custom}
\bibliographystyle{acl_natbib}

\clearpage
\appendix
\begin{table*}[t]
    \small
    \centering
    {
    \begin{tabular}{l|c|c|c|c|c}
   
        \toprule
         
        Model & \textsc{PerpInd} & \textsc{PerpOrg}  & \textsc{Target} & \textsc{Victim} & \textsc{Weapon} \\ 
        \midrule
        
         \dygiepp~  & \multirow{2}{*}{48.39/ 32.61/ 43.32} & \multirow{2}{*}{56.00/ 34.15/ 42.42} & \multirow{2}{*}{53.49/ 50.74/ 52.08} & \multirow{2}{*}{60.00/ 66.32/ 63.00} & \multirow{2}{*}{57.14/ 53.33/ 55.17}\\
         \cite{wadden-etal-2019-entity}\\
         
         GRIT  & \multirow{2}{*}{65.48/ 39.86/ \textbf{49.55}} & \multirow{2}{*}{66.04/ 42.68/ 51.85} & \multirow{2}{*}{55.05/ 44.12 / 48.98} & \multirow{2}{*}{76.32/ 61.05/ 67.84} & \multirow{2}{*}{61.82/ 56.67 / 59.13}\\
         \cite{du-2020-grit}\\
        \midrule
        \modelshort~ & 67.12/ 35.51/ 46.45 & 67.12/ 59.76/ \textbf{63.23} & 64.13/ 43.38/ \textbf{51.75} & 77.22/ 64.21/ \textbf{70.11} & 67.27/ 61.67/ \textbf{64.35} \\
        
        \bottomrule
    \end{tabular}
    }
    \vspace{-2mm}
    \caption{Performance breakdown with regard to each role in CEAF-REE (Precision/ Recall /F1) on the MUC-4 REE task.}
    \label{tab:ree_breakdown}
    
\end{table*}
\begin{table*}[h]
    \small
    \centering
    {
    \begin{tabular}{lccccccccc}
        \toprule
        
        & \multicolumn{3}{c}{\textbf{Role-filler Entity Extraction}} & 
        \multicolumn{3}{c}{\textbf{Binary Relation Extraction}} & 
        \multicolumn{3}{c}{\textbf{4-ary Relation Extraction}} \\ \midrule
        Model      & Precision & Recall &  F1 &  Precision & Recall &  F1 & Precision & Recall &  F1    \\ 
        \midrule
        TANL & 58.42 & \textbf{46.74} & 51.93 & 3.12 & 2.11 & 2.39 & 0.00 & 0.00 & 0.00 \\
        \modelshort~&  \textbf{61.34} & 46.11 & \textbf{52.64} & \textbf{22.04} & \textbf{19.24} & \textbf{19.60} & \textbf{1.38} & \textbf{2.77} & \textbf{1.85} \\

        \bottomrule
    \end{tabular}
    }
    \vspace{-2mm}
    \caption{Corresponding development set performance of the reported test set results in \Cref{tab:main}.}
    \label{tab:dev_set_performance}
    
\end{table*}
% \section{Implicit information}
% \Steeve{Give examples of implicit information in SicREX and MUC.}

\section{REE Performance Breakdown}
\label{sec:ree_breakdown}
\Cref{tab:ree_breakdown} demonstrates the per-role performance comparison between \modelshort~ and other baselines. We observe that:
\begin{itemize}
\item \modelshort~ achieves the best precision across all roles.
\item Except for \textsc{PerpInd}, \modelshort~ obtain substantial improvement in F1 over other baselines.
\item While \modelshort~ has higher precision over GRIT in extracting \textsc{PerpInd} entities, it scores slightly lower in recall, leading to worse F1 performance.

\end{itemize}

\section{TANL Decoding Target Formulation}
\label{sec:TANL_targets}
In this section, we illustrate how we formulate the TANL \cite{paolini2021structured} decoding targets for REE and RE. The formulation for REE is simple due to its similarity to the NER task. We produce REE decoding targets exactly the same way as how NER decoding targets are formed in \citet{paolini2021structured}. Given the REE example in \Cref{fig:method}, the corresponding TANL decoding target is:
\begin{quote}
    Two U.S. mormon missionaries -- aged 19 and 21 -- were shot to death last night by [a group of terrorists| \textit{PerpInd}] from the [Zarate Wilka Armed Forces of Liberation| \textit{PerpOrg}] (FAL) ... blew up the [lines| \textit{Target}] providing power to La Paz, ... the U.S. citizens -- [Todd Ray Wilson Burdenson| \textit{Victim}] and Jeffrey Brent Ball -- ... they were killed with two bursts of [machinegun| \textit{Weapon}] fire ...
\end{quote}

As for RE, we follow how \citet{paolini2021structured} handles \textit{nested entities and multiple relations}, but we made a small modification on decoding targets. Since \textsc{SciREX} does not contain relation type annotation, we use the related entities' types as the relation type in the decoding targets. With their formulation, the decoding target is created by inserting each relation annotation around the first-appearing entity in the input document. Take the RE instance in \Cref{fig:method} as an example. The corresponding TANL decoding target would be:

\begin{quote}
    Introduction: [Natural language inference | \textit{Task} | \textit{Method} = aESIM] (NLI) is an important andsignificant task in natural language processing (NLP)... \\
    Method: We ... denote the modified ESIM as aESIM ...\\
    Experiments: The [accuracy | \textit{Metric} | \textit{Material} = Quora] (ACC) of each method is measured by thecommonly used precision score ... It also achieved 88.01 \% on Quora ...
\end{quote}

\section{Hardware and Software configurations}
All experiments are conducted on a CentOS Linux 7 (Core) machine with NVIDIA RTX 2080. We use PyTorch 1.6.0 with CUDA 10.1 as the Deep Learning framework and utilize Transformers 4.3.0 to load all pre-trained language models. 

\section{Implementation Details}
We conducted grid search to find the best learning rate over $\{1\times 10^{-5}, 3\times 10^{-5}, 5\times 10^{-5}, 7\times 10^{-5}, 9\times 10^{-5}\}$ using \textbf{\modelshort~ w/o \topkcopy~} on the MUC-4 REE task. The best learning rate, $5\times 10^{-5}$, is fixed for all other experiments. Models are trained for 150 epochs for REE and binary RE experiments, and 50 epochs for 4-ary RE experiments. To reproduce our results, please follow the README.md file in \url{https://github.com/PlusLabNLP/TempGen}. The weights of the trained models are also included for reproduction purposes.

\section{Validation Performance}
For all reported test set results in \Cref{tab:main}, the corresponding development set performance are listed in \Cref{tab:dev_set_performance}.

\end{document}